\documentclass{article}

\usepackage[preprint]{neurips_2026}

\usepackage[utf8]{inputenc} %
\usepackage[T1]{fontenc}    %
\usepackage{hyperref}       %
\usepackage{url}            %
\usepackage{booktabs}       %
\usepackage{amsfonts}       %
\usepackage{nicefrac}       %
\usepackage{microtype}      %
\usepackage{xcolor}         %

\usepackage{amsmath}
\usepackage{amssymb}
\usepackage{mathtools}
\usepackage{amsthm}
\usepackage{placeins}

\usepackage[capitalize,noabbrev]{cleveref}

\theoremstyle{plain}

\theoremstyle{definition}

\theoremstyle{remark}

\usepackage[textsize=tiny]{todonotes}

\usepackage[most]{tcolorbox}
\tcbuselibrary{listings,minted,skins,breakable}

\setminted{
  breaklines,
  breakanywhere,         %
  autogobble,            %
  tabsize=2,
  fontsize=\footnotesize,
  encoding=utf8,
  texcomments=false,
  linenos,
  numbersep=6pt,
  breaksymbol=\raisebox{0.5ex}{\tiny\ensuremath{\hookrightarrow}},
}

\tcbset{
  promptbox/.style={
    enhanced,
    breakable,
    sharp corners,
    boxrule=0.4pt,
    colback=black!1,
    colframe=black!15,
    coltitle=black,
    left=6pt,right=6pt,top=6pt,bottom=6pt,
    attach boxed title to top left={yshift=-2pt, xshift=2pt},
    boxed title style={
      size=small,
      colback=black!3,
      colframe=black!15,
      sharp corners
    },
  }
}

\tcbset{
  lang/.style      = { minted language=#1 },               %
  pygstyle/.style  = { minted options={style=#1} },        %
  nolinenos/.style = { minted options={linenos=false} },   %
  tinytext/.style  = { minted options={fontsize=\scriptsize} },
}

\newtcblisting{promptblock}[2][]{%
  promptbox,
  listing engine=minted,
  minted language=markdown, %
  minted options={},
  listing only,
  title={#2},
  #1
}

\newtcbinputlisting{\includeprompt}[3][]{%
  promptbox,
  listing engine=minted,
  minted language=markdown, %
  minted options={},
  listing only,
  title={#2},
  listing file={#3},
  #1
}

\definecolor{questioncolor}{RGB}{45, 52, 54}    %
\definecolor{answercolor}{RGB}{99, 110, 114}    %
\definecolor{tokencolor}{RGB}{52, 152, 219}     %
\definecolor{responsecolor}{RGB}{47, 54, 64}    %

\definecolor{gentokens}{RGB}{120, 128, 250}

\newtcolorbox{modelresponsevar}{
  enhanced,
  colback=white,
  colframe=questioncolor,
  arc=0mm,
  boxrule=0.5pt,
  fonttitle=\bfseries,
  coltitle=white,
  colbacktitle=questioncolor,
  title=Example,
  attach boxed title to top left={yshift=-\tcboxedtitleheight/2},
  boxed title style={sharp corners, boxrule=0pt},
  width=\linewidth
}

\newcommand{\modelexamplecorrect}[5]{%
  \begin{modelresponsevar}%
    \begin{tcolorbox}[colback=white!95!questioncolor,colframe=questioncolor,title=Question,fonttitle=\bfseries]
      #1
    \end{tcolorbox}

    \begin{tcolorbox}[colback=white!95!tokencolor,colframe=tokencolor,title=Effort Percentage Specified,fonttitle=\bfseries]
      #2
    \end{tcolorbox}
    
    \begin{tcolorbox}[colback=white!98!responsecolor,colframe=responsecolor,title=Model Response,fonttitle=\bfseries,breakable]
      #3
    \end{tcolorbox}

    \begin{tcolorbox}[colback=white!95!questioncolor,colframe=questioncolor,title=Correct Answer,fonttitle=\bfseries]
      #4
    \end{tcolorbox}

    \begin{tcolorbox}[colback=white!95!gentokens,colframe=gentokens,title=Generated Tokens,fonttitle=\bfseries]
      #5
    \end{tcolorbox}
    \end{modelresponsevar}%

}

\title{e1: Learning Adaptive Control of Reasoning Effort}

\author{
Michael Kleinman\thanks{Correspondence to Michael Kleinman \texttt{<mjklein@amazon.com>}}, Matthew Trager, Alessandro Achille, Wei Xia, Stefano Soatto \\
AWS Agentic AI
}

\begin{document}

\maketitle

\begin{abstract}
Increasing the thinking budget of AI models can significantly improve accuracy, but not all questions warrant the same amount of reasoning. Users may prefer to allocate different amounts of reasoning effort depending on how they value output quality versus latency and cost.
To leverage this tradeoff effectively, users need fine-grained control over the amount of thinking used for a particular query, but few approaches enable such control.
Existing methods require users to specify the absolute number of desired tokens, but this requires knowing the difficulty of the problem beforehand to appropriately set the token budget for a query.
To address these issues, we propose \textit{Adaptive Effort Control}, a self-adaptive reinforcement learning method that trains models to use a user-specified fraction of tokens relative to the current average chain-of-thought length for each query. This approach eliminates dataset- and phase-specific hyperparameter tuning. 
Users can dynamically adjust the cost-accuracy trade-off through a continuous effort parameter specified at inference time.
We observe that the model automatically learns to allocate resources proportionally to the task difficulty. Across model scales ranging from 1.5B to 32B parameters, our approach reduces chain-of-thought length on math tasks  by 1.9-3.7x at matched accuracy relative to the base model used for RL training. 
\end{abstract}

\section{Introduction}

Models can improve accuracy by increasing the amount of test-time compute, \textit{i.e.,} the length of chain-of-thought.
In fact, with unbounded test-time compute, many problems can be trivially solved through brute force search \citep{achille2025ai}. However, this would result in untenable compute cost and latency. 
Instead, AI agents should learn to allocate their compute budget to achieve the best trade-off between cost/latency and accuracy, which depends on the task, the user, and the environment: When compute is cheap, producing more tokens may be desirable. When the user assigns low economic value to a task, compute usage should be proportionally lower, lest the user spend more than they gain.

Optimization of this tradeoff can be framed as optimizing the expected \textit{net reward}:
\begin{equation*}
J = E[R - \lambda T]  
\end{equation*}
where $R$ is the task reward, $T$ is the time used (number of tokens), and $\lambda$ is the cost per token (which depends on the model and infrastructure). The user then selects $\lambda$ based on the perceived value of the task, and the accuracy trade-off they want to achieve. 

From an economic perspective, $J$ is the right measure to evaluate model performance. However, directly using the net reward for RL training creates issues: when $\lambda$ is fixed and the reward is not calibrated to task difficulty, complex tasks requiring many tokens are disproportionately penalized. 
Recent work has used a modified penalty for \mbox{token usage \citep{arora2025training} or adaptive} values of $\lambda$ \citep{xiang2025just}, however these approaches do not enable inference-time control over the compute-accuracy tradeoff. 

To enable inference-time control and obtain strong accuracy-versus-cost tradeoffs across tasks of varying difficulty, we introduce \textbf{Adaptive Effort Control} (AEC), a simple auto-adaptive training method that teaches models to balance solution quality and reasoning cost according to user control. %
Rather than asking the model to achieve an absolute trade-off controlled by $\lambda$, we ask the model to produce an answer using less than $r$ times the average amount of tokens used by correct solutions to the problem so far:
\begin{equation}
\begin{aligned}
\max R \quad \text{ s.t. } \quad \frac{T}{T_\text{avg}} < r.
\end{aligned}
\end{equation}
Reducing $r$ 
forces the model to learn cheaper solutions, %
without requiring a $\lambda$ tuned to the particular task. Since time is normalized by the average length of a correct solution, equal effort is devoted to optimizing both easy and difficult problems. 
If desired, the resulting model can then be calibrated to take $\lambda$ as input, or any other user-facing control-knob for effort allocation. Rather than fixing the hyperparameter $r$, we specify it in the prompt on a per-query basis during training, and use a reward function that depends on the value of $r$. Our reward function integrates directly with standard RL training procedures like GRPO \citep{shao2024deepseekmath}.

After training with \textbf{AEC}, on our resulting model \texttt{e1}, increasing the effort parameter $r$ leads to an overall increase in both the number of generated tokens and task performance. We observe this across a range of tasks that require varying number of tokens.  

Additionally, across model scales from 1.5B to 32B parameters, our approach enables significant efficiency gains. On math tasks, we find that our training approach results in a 1.9-3.7x reduction in chain-of-thought length at matched accuracy relative to the base model used for our RL training. Furthermore, after training, it is possible to calibrate the effort parameter to provide linear control over either relative accuracy or relative token usage, enabling intuitive user control tailored to different applications.

\section{Related Work}

\paragraph{Test-Time Scaling for LLMs.} 
Several works have shown that LLM performance can be improved by increasing computation at inference time in different ways. Methods can involve repeated sampling \citep{lightman2023let, brown2024large}, search-based techniques with a verifier \citep{wu2024inference, snell2024scaling, manvi2024adaptive, uscidda2025latts}, or increasing the ``reasoning,'' or chain-of-thought computation, before producing an answer \citep{guo2025deepseek, team2025kimi}. However, increased accuracy does not necessarily lead to more ``intelligent'' behavior \citep{achille2025ai}, as the models can simply improve accuracy by using the additional resources for extended brute force search, without learning new strategies. 
On the other hand, learning to minimize inference time to find a correct solution leads to maximizing the \textit{algorithmic information} in the trained model \citep{achille2025ai}. 

\paragraph{Efficient Reasoning Models.} While accuracy typically improves with longer chain-of-thought reasoning, longer responses incur higher latency and computational costs.  As a result, not only is more compute leading to diminishing return, but to decreasing net utility. For example, a model that improves performance in a math test but returns every answer after the time limit has passed is worth less than a model that guesses the answer, \textit{i.e.,} performs at chance level.  A recent class of methods aim for efficient reasoning \citep{arora2025training, zhang2025grpo} using a penalty for the length of the  response. \citep{xiang2025just} use an adaptive length penalty depending on pass-rate of the model on the problem.
These works typically use a fixed hyperparameter which leads to a single operating point along an accuracy-versus-tokens curve. In contrast, in our work we 
specify a control input along with the query, which allows us to trace the entire cost-control curve
of the trained model 
and therefore operate at any point along the curve.

\paragraph{Length-Control for Reasoning Models.}  
To enable control over the generation length, \cite{muennighoff2025s1} proposed the S1 method
which modifies inference by either forcing early termination at a specified number of tokens, or forcing a continuation if the model wanted to terminate early. %
Similarly, \cite{aggarwal2025l1} train models to use either an exact or maximum number of tokens by conditioning on a target length in the prompt and modifying the corresponding reward function accordingly. However, such approaches require users to know the difficulty of the task beforehand to appropriately set the number of tokens.
Instead of fixing a hard thinking budget, concurrent works propose training model with soft prompts requesting ``low'', ``medium'', or ``high'' thinking effort \citep{agarwal2025gpt, he2025thinkdial}. In particular, \cite{he2025thinkdial} presents a recipe to train models to reason at such effort levels though their approach involves multiple training stages, relies on custom system prompts per effort level, and doesn’t allow fine-grained control.

\section{Method}

\begin{figure*}[t]
    \centering
\includegraphics[width=0.4\linewidth]{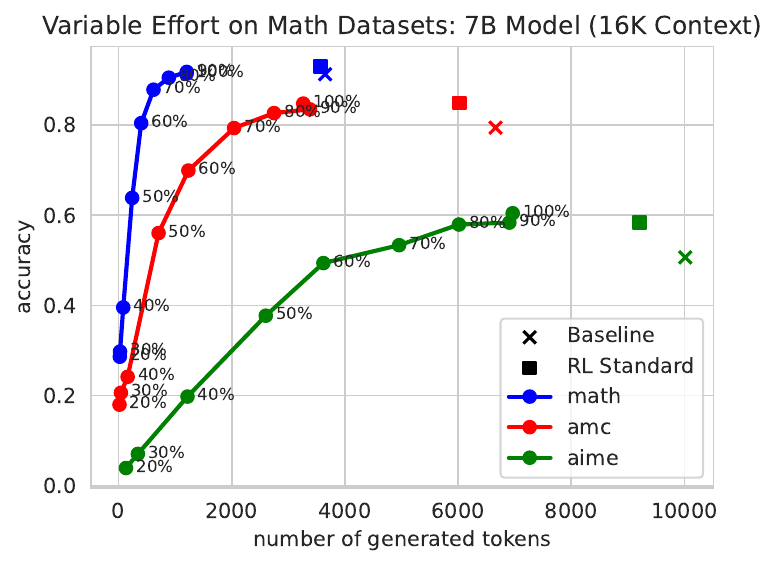}
    \includegraphics[width=0.4\linewidth]{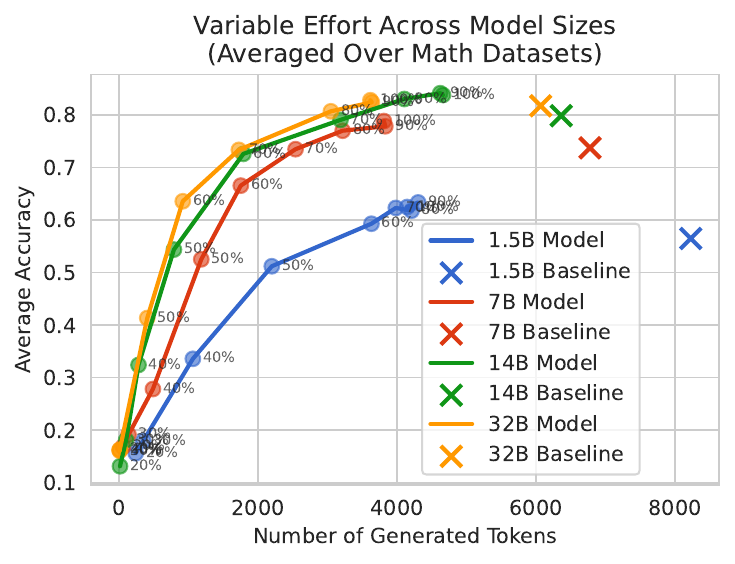}
    \vspace{-0.5em}
    \caption{\textbf{Variable Effort Control and Improved Reasoning Efficiency. }\textbf{(Left)}  
    After training, increasing $r$ in the prompt leads to an overall increase in both accuracy and the number of generated tokens across datasets of varying difficulty. 
    Note that in addition to allowing control, this training approach also allows us to significantly reduce the number of generated tokens compared to the baseline (in crosses) and with regular RL training (squares).
    \textbf{(Right)} We apply our approach across model scales ranging from 1.5B to 32B and show results averaged across math datasets (see Appendix Fig.~\ref{fig:app-model-sizes} for additional dataset-specific results), with our approach enabling more efficient reasoning and improved performance relative to the base models. We observe superior performance vs. number of generated token curves with increasing model size. }
    \vspace{-1em}
    \label{fig:main_math_and_ood}
\end{figure*}

Our goal is to train a model that can effectively balance 
solution quality and reasoning cost on a per-query basis according 
to user-specified preferences. 
We introduce \textit{Adaptive Effort Control} (AEC), a training method that automatically adapts the target effort level for the task to both the complexity of the task and the current status of training. Given a training example $x$ and a sampled target \textit{effort ratio} $r > 0$, we proceed as follows. First, we construct an input $x_r$ that encoded both the task $x$ and the target effort-ratio $r$. In our experiments, we simply concatenate $r$ to the prompt as follows:
\begin{equation*}
x_r = \text{Concat}(x, \ \text{``Let's spend $\{r \cdot 100 \}\%$ effort.''} )
\end{equation*}
Given $x_r$, we sample $N$ chain-of-thought traces ($N=16$ in our experiments) and denote with $S = \{h_1, ..., h_{N_S(x_r)}\}$ the set of traces that terminate with a correct solution. 
We write $\ell(h)$ for the total number of tokens generated in a trace $h$, which we refer to interchangeably as the chain-of-thought length.
We define the average time to find a solution to $x_r$ as the average length of the successful traces:
\[
T_\text{avg}(x_r) = \frac{\alpha}{N_S(x_r)} \sum_{i=1}^{N_S(x_r)} \ell(h_i),
\]
or $T_\text{avg}(x_r) = \infty$ if no correct solution was found. $\alpha = 2.5$ is a fixed scalar across all examples allowing $r \in [r_{min}, 1]$ to vary in a more interpretable range when prompting the model.
We then define the \textit{AEC reward} for each trace $h$ as:
\begin{equation}
    \label{eq:main-objective}
    \boxed{
    R_\text{AEC} \big(y, y^*|x, r \big) = R \big(y, y^*|x \big) \cdot \mathbb{I}\left(\frac{\ell(h)}{T_\text{avg}(x_r)} < r \right)}.
\end{equation}
That is, the reward function gives no reward to any trace that uses more than a fraction $r$ of the average time required to solve the problem. Crucially, since the constraint is relative to the \textit{current} average time required to solve the \textit{specific task} $x$, the AEC reward automatically adapts both to the complexity of the problem and the current stage during training.
If no continuations are correct, all rollouts receive a reward 0.

\begin{figure}[t]
    \centering
    \includegraphics[width=0.48\linewidth]{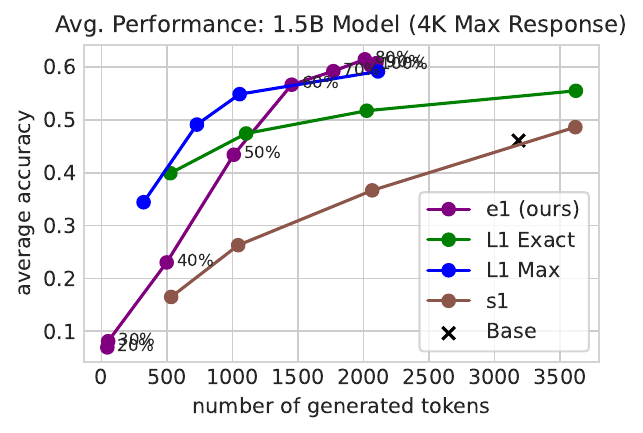}
    \includegraphics[width=0.48\linewidth]{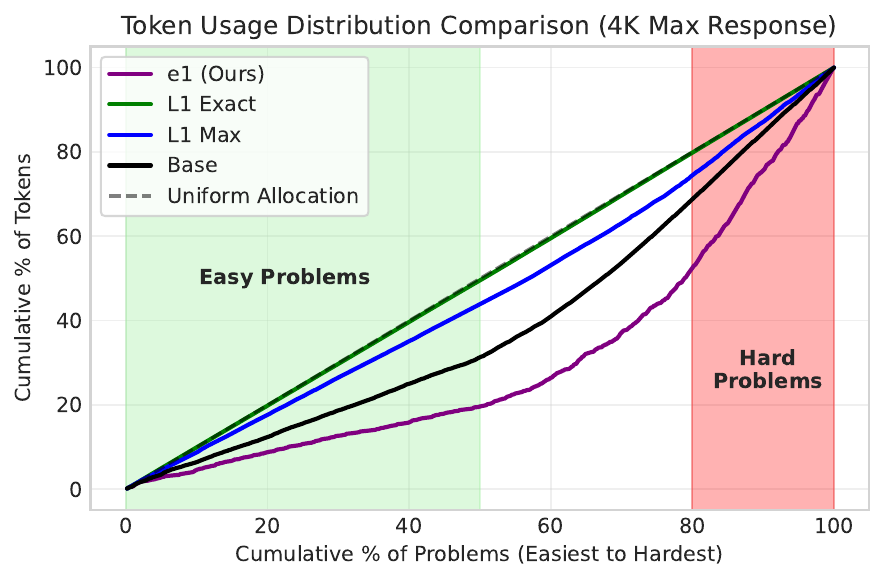}
    \vspace{-0.5em}
    \caption{\textbf{(Left) Comparison against other length-control baselines.} 
    Our controllable model (\texttt{e1}) obtains a superior accuracy-token tradeoff compared to S1 across all effort levels.
    When the effort level is above 60\%, it also outperforms L1-Exact, and slightly outperforms L1-Max (which does not allow precise control over the generation length).
    See Appendix Fig.~\ref{fig:app-l1-comparison} for performance on individual datasets. \textbf{(Right) Adaptation to query difficulty.} 
    In addition to providing control over the number of tokens, \texttt{e1} is automatically able to adapt the length of the reasoning chain to  the difficulty of the input task.
    We see this by ordering the problems from easiest to hardest (based on fraction of times the model solves a problem correctly) and plotting the cumulative percentage of total tokens used for solving those problems. A diagonal line indicates equal tokens allocation to each problem, and increasingly convex curves indicate fewer token allocation to easier problems.
    Our approach (shown for $r=1$, purple) allocates a smaller percentage of tokens to easier problems than other length-control baselines and the base model used for RL training. In particular our model allocates only $\sim$20\% of tokens to the easiest 50\% of problems. See Appendix Fig.~\ref{fig:difficulty-adaptation} for analysis of the 7B model's adaptation to query difficulty.
    }
    \vspace{-0.5em}
    \label{fig:l1-comparison}
\end{figure}

\paragraph{Training Algorithm.} %
We trained our model \texttt{e1} using our AEC reward function (Eq.~\ref{eq:main-objective}). For optimization we used GRPO \citep{shao2024deepseekmath} without any modifications. We use binary rewards for answer correctness such that $R(y, y^*|x) = \mathbb{I}(y = y^*)$. We refer to our training procedure as \emph{AEC training}.%

\paragraph{Training data.} We create our training dataset as follows. We start with a training dataset $\mathcal{D} = \{(x^i, y^{i*})\}_{i=1}^N$ for reinforcement learning where $x^i$ denotes the input question, and $y^{i*}$ denotes the corresponding answer.
For each question $x^i$ we sample $r \in [r_{min}, 1]$. This augmentation results in a new training dataset of $\mathcal{D}^{new} = \{(x_{r}^i, y^{i*})\}_{i=1}^N$ that we use for our reinforcement learning training. At inference time, we augment the questions in a test dataset with a particular effort level $r$ of interest and quantify the ability to control the number of generated tokens and performance.

\section{Experimental Setup}

\label{sec:exp-setup}

\paragraph{Models and Datasets.} 
We start with an R1-Distilled Qwen \{1.5,7,14,32\}B as the base model, unless otherwise stated. This is a distilled version of the DeepSeek R1 model \citep{guo2025deepseek} onto the Qwen 2.5 family of models \citep{qwen2025qwen25technicalreport}. This model had strong mathematical capabilities but does not have the ability to modulate its thinking effort. In the main text, we primarily report results on the 7B model and state when other models are used. We train using the DeepScaler dataset \citep{luo2025deepscaler} which consists of $\sim$40,000 math problems consisting of question and ground-truth output. We focus our evaluations on a variety of math datasets of varying difficulty (AIME 2024, AMC, and MATH500 \citep{hendrycks2021measuring}) and also evaluate the ability to transfer to out-of-domain datasets. We also evaluate the model out-of-domain using GPQA Diamond \citep{rein2024gpqa}, LSAT \citep{zhong2023agieval}, and MMLU \citep{hendrycks2020measuring}; for MMLU we use a random subset of 1,000 questions.

\paragraph{Hyperparameters and Training Details.} We train and evaluate with the relative effort $r \in [0.2,1]$.
We set $\alpha = 2.5$. 
Unless explicitly stated, we trained with a context length of 16K tokens, of which 15K were allocated for the response. At evaluation, the maximum response length was kept the same. 
For training, we use a batch size of $320$, and a learning rate of 1e-6. We train for $1000$ steps unless otherwise stated. We trained with GRPO using 16 rollouts per question. We used the VERL framework for training \citep{sheng2024hybridflow}. 

\paragraph{Evaluation.} We evaluate the ability for our approach to modulate both the performance and number of generated tokens. We assess this by plotting the accuracy against number of generated tokens as we vary the effort level parameter $r$ across the datasets described above. When doing evaluations, we sample $16$ times per question and report the average accuracy (fraction of questions solved correctly).
We evaluate by augmenting the evaluation dataset using with the effort $r \in [0.2, 0.3, \ldots, 1]$.

\paragraph{Baselines.} We compare our approach against length-control baselines L1 \citep{aggarwal2025l1} and S1 \cite{muennighoff2025s1}. S1 modifies inference to extend inference by appending \texttt{"Wait"} if the model wanted to conclude before the user's targeted number of tokens, or by forcing termination at the user's specified number of tokens. In our experiments with S1, we enforce the token specification by appending \texttt{"Wait"} up to $100$ times. L1 trains a model using reinforcement learning to output an answer while following a desired number of tokens specified in the prompt. L1 consists of two variants: L1-Exact aims to exactly follow of the number of tokens specified by a  user, whereas L1-Max aims to satisfy a maximum number of tokens constraint. %

\section{Results}

\subsection{Variable Effort Control and Improved Reasoning Efficiency}

We first assess whether our approach enables variable effort control on math evaluation datasets. In Fig.~\ref{fig:main_math_and_ood} (Left) we show that after training, increasing the effort $r$ in the prompt leads to an overall increase in accuracy and number of generated tokens across datasets of varying difficulty.
We find that the model uses more tokens on more difficult problems, for example allocating more budget for problems in the AIME dataset which is generally considered more difficult than the MATH500 dataset, where our model allocates less budget.
In addition to allowing control, our training approach leads to significant reductions in the number of generated tokens compared to the base model used for RL training (3x reduction at equal performance, averaged over tasks) and models trained with standard reinforcement learning, while allowing for increasing performance with more generated tokens.

In Fig.~\ref{fig:main_math_and_ood} (Right) we show that our approach can be applied to models of various sizes (ranging from 1.5B to 32B), enabling more efficient reasoning while maintaining or improving performance relative to the base model used for our RL training. We observe better performance vs. number of generated token curves with increasing model size. We show aggregated results in Fig.~\ref{fig:main_math_and_ood} (Right), and we show dataset-specific results for the 1.5B, 14B, and 32B models in Appendix Fig.~\ref{fig:app-model-sizes}. 
Across different math tasks and model sizes, we observe a 1.9-3.7x reduction in generated tokens when matching the baseline performance level, with larger reductions for smaller models (Appendix Table~\ref{tab:reductions}). We also apply AEC to an instruction-tuned model and find that accuracy and number of generated tokens increase overall with the effort level (Appendix Fig.~\ref{fig:app-effect-base}).

\subsection{Adaptation to query difficulty and comparison to length-control baselines} 

In addition to providing a new mechanism for controlling the length of generation, we show that our approach has strong performance-cost tradeoffs compared to other token-based specification methods (Fig.~\ref{fig:l1-comparison}, Left). We do this ablation on 1.5B models trained with a smaller maximum response length (4K tokens) using the DeepScaler model \citep{luo2025deepscaler} as the base model to align with previous work and allow us to leverage available online checkpoints for comparison \citep{aggarwal2025l1}. Additionally, we show that training with our approach leads to smaller token allocation to easier problems, and larger token allocation to more difficult problems (Fig.~\ref{fig:l1-comparison}, Right). Our approach (shown for maximum effort $r = 1$) allocates fewer tokens to easier problems than uniform allocation and the baseline model, allocating $\sim$20\% of tokens to 50\% of problems. This differs from exact token-based specification which uses the specified number of tokens regardless of question difficulty.

\vspace{-0.3em}
\subsection{Calibration for interpretable control over percentage effort} 
\vspace{-0.3em}
We saw previously that our approach enables increasing tokens and performance with increasing $r$, but it was not precisely clear how changing $r$ affects the number of generated tokens or accuracy. We show that a simple reparameterization of the effort parameter after training can enable linear control for either relative tokens or relative accuracy (defined with respect to the tokens or accuracy with $r=1$). %

\begin{figure}[t]
    \centering    \includegraphics[width=0.35\linewidth]{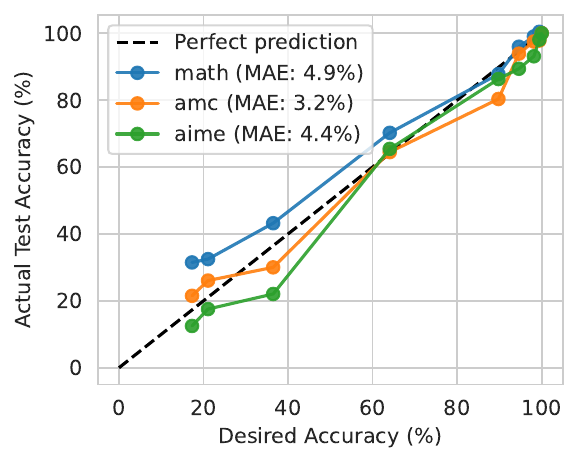}
    \includegraphics[width=0.35\linewidth]{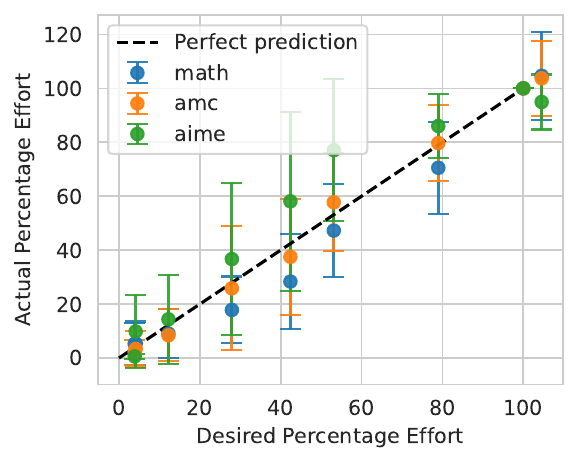}
    \caption{\textbf{Calibration for interpretable control over percentage effort.} After training, we can reparameterize the effort parameter $r$ to linearly control either the \textbf{(Left)} relative accuracy or  \textbf{(Right)} relative number of tokens and enable more intuitive user control. Error bars denote standard deviation.
    }
    \vspace{-0.5em}
    \label{fig:dynamics-calibration}
\end{figure}

When calibrating for the relative number of tokens, we define the maximum effort for a problem $x$ to be the average number of tokens generated $T(x,r=1)$ when using the largest value of $r$.
On a small validation set (15 examples from each of MATH500, AMC and AIME) we compute the desired percentage effort as $d(r) = \mathbb{E}_x \left[\frac{T(x,r)}{T(x,r=1)}\right]$, allowing us to map from user-specified desired percentage effort  $d$ to $r$ (that is used to augment the prompt). On new examples, we similarly compute the outputted percentage effort as $o(x,r) = \frac{T(x,r)}{T(x,r=1)}$. We find that the outputted percentage effort is highly correlated with the desired percentage effort (Fig.~\ref{fig:dynamics-calibration}, Right) even though the absolute number of tokens differs significantly across these datasets. We summarize these results by computing the average relative error $e = \mathbb{E}_{x \in D, r} [o(x,r) - d(r)]$ over the different datasets $D$, finding that the mean average relative error across datasets is 11\%  (Fig.~\ref{fig:app-calibration}). 

We similarly calibrate based on accuracy by computing $a(r) = \frac{\text{Acc(r)}}{Acc(r=1)}$ using the same validation set. We compare the desired accuracy $a(r)$ that a user can specify against the accuracy on the test set in Fig.~\ref{fig:dynamics-calibration} (Left), finding a mean absolute error (MAE) of 4\% across effort levels and datasets. 

\subsection{Learning dynamics of AEC training}

\begin{figure*}[t]
    \centering
    \includegraphics[width=0.8\linewidth]{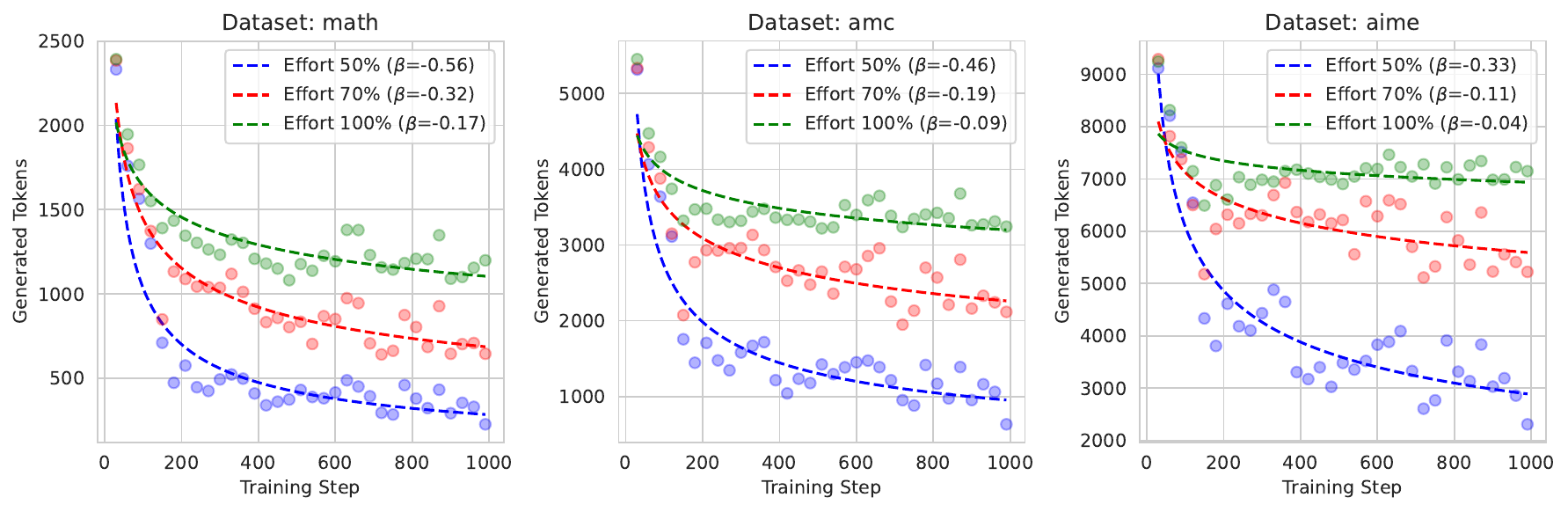}
    \includegraphics[width=0.8\linewidth]{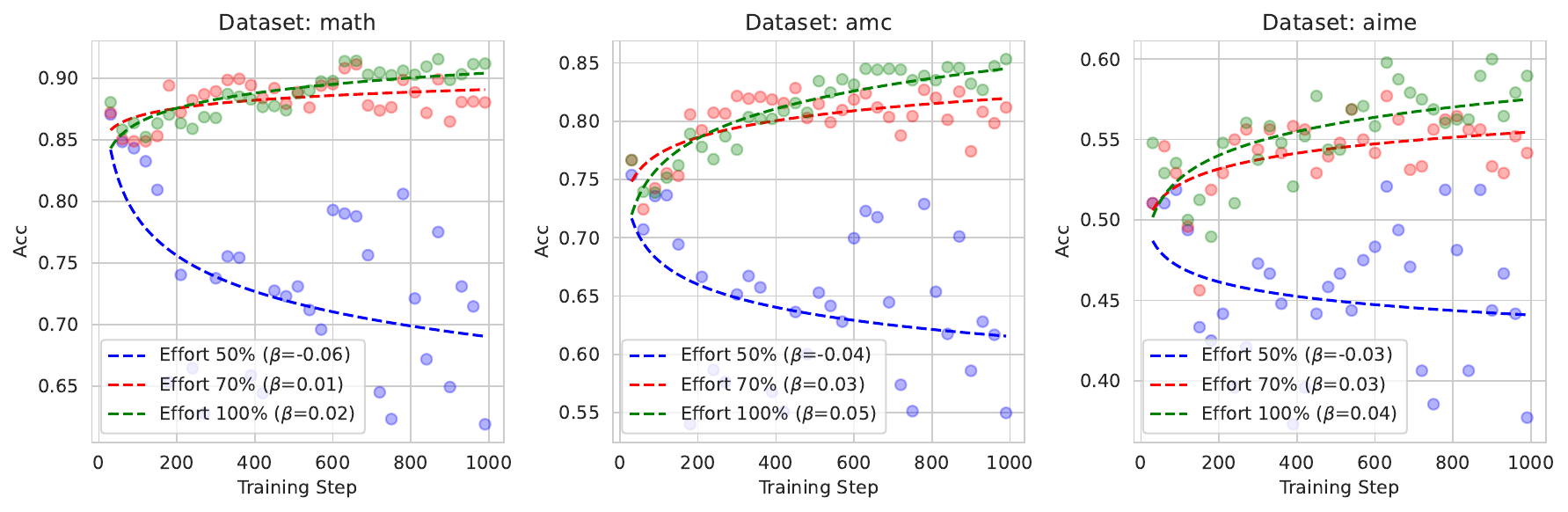}
    \vspace{-0.5em}
    \caption{
    \textbf{Learning dynamics of AEC training.} 
We plot the accuracy and number of tokens as a function of training step for the three math evaluation datasets. Early in training, the model has similar accuracy and number of generated tokens regardless of the effort level, which means that the model needed to learn to become sensitive to the effort level. We find that the number generated tokens roughly decreases as a power law with exponent $\beta$ across training steps. We find that shorter problems (MATH500) decrease faster (larger $\beta$) than longer problems (AIME). Despite token lengths decreasing through training, we find that accuracy increases when the effort level is at least $70\%$.
    }
    \label{fig:learning-dynamics}
\end{figure*}

We have observed that after our reinforcement learning training, specifying the effort level in the prompt enables varying the accuracy and number of tokens on a per-question basis. But how does this ability emerge through the RL training?

We analyze how the accuracy and number of generated tokens vary over learning (Fig~\ref{fig:learning-dynamics}). 
We plot the accuracy and number of tokens as a function of training step for the three math evaluation datasets.
At the beginning of training, the number of tokens and accuracy was similar regardless of the effort level supplied in the prompt.
We find that the number of tokens roughly decreases as a power law. We find that shorter problems (MATH500) decrease faster than longer problems (AIME).
Despite token lengths decreasing through training, we find that accuracy increases when the effort level is at least $70\%$. This highlights that in addition to enabling controllable inference-time performance, our training enables both more efficient responses and performance increase. We observe similar learning dynamics across models of various sizes (Appendix Fig.~\ref{fig:learning-dynamics-32b}, Fig.~\ref{fig:learning-dynamics-14b}, Fig.~\ref{fig:learning-dynamics-1.5b}).

In Fig.~\ref{fig:relative-toks-acc}, we observe that the relative number or tokens (or accuracy) compared to $100$\% effort is consistent across datasets throughout training for various effort level parameters $r$.

\subsection{Transferability of effort control across different domains}

Despite training on math questions, we find that after AEC training, we can modulate the number of generated tokens and performance on  datasets in different domains by varying the effort parameter $r$ (Fig.~\ref{fig:ood_and_effect-objective}, Left). We observe improved performance on LSAT with reduced number of generated tokens compared to the model before our RL training (on math questions), and observe comparable performance to a model trained using standard RL training, while using fewer tokens. %
We observe reduced performance on GPQA Diamond and MMLU compared to the baseline and an RL trained model, though our approach uses significantly fewer tokens. Importantly, across all of the datasets we find that the number of generated tokens and accuracy increase overall with the effort parameter supplied in the prompt.

\subsection{Effect of reward function and robustness to prompt}

Here, we aim to ablate what aspects of our reward function resulted in the ability for inference-time control using the effort parameter. We examine the effect of the prompt and the reward function, and show this ablation for a 1.5B model. For the following experiments we use a different prompt:
\begin{equation*}
\begin{split}
x_i^{new} = \text{``(\{$p$\} \texttt{pts}) $\{x_i\}$ We'll get \{$p$\} points for correctly answering the question.''}
\end{split}
\end{equation*}

\begin{figure}[t]
    \centering
    \includegraphics[width=0.4\linewidth]{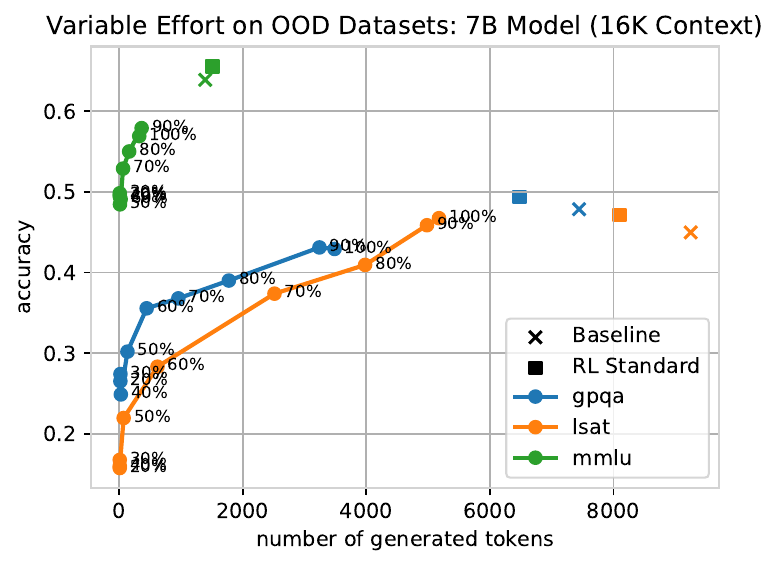}
    \includegraphics[width=0.47\linewidth]{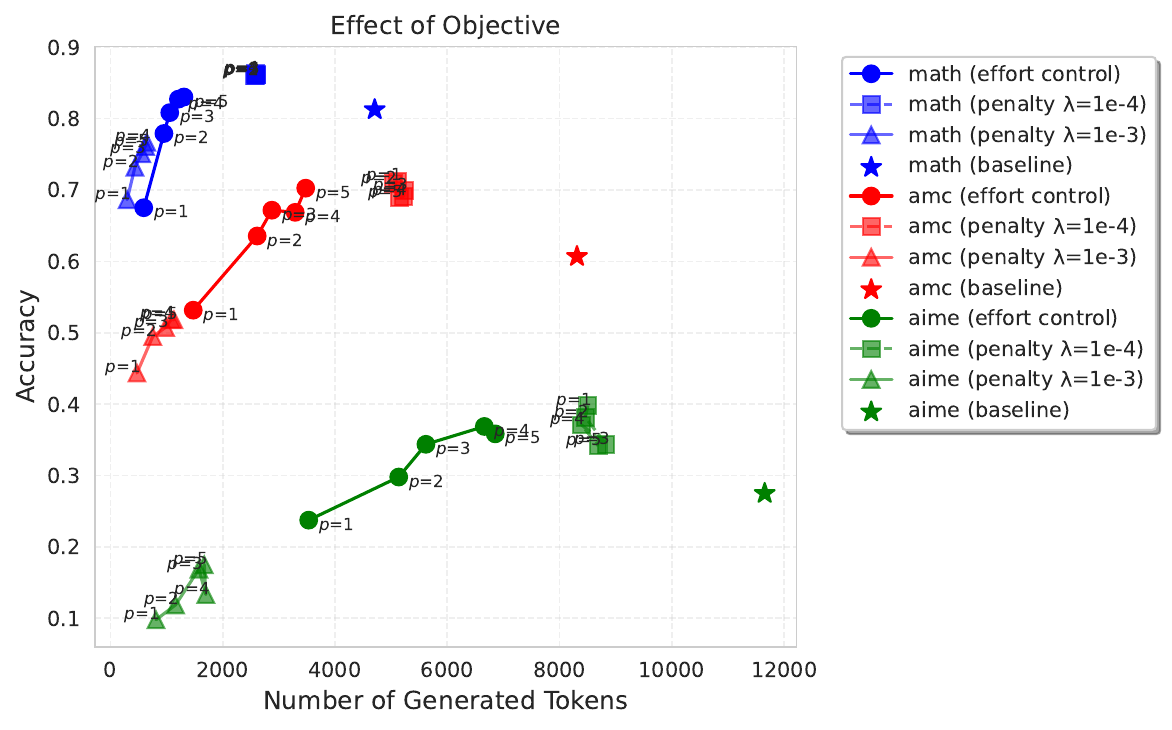}
    \vspace{-0.5em}
    \caption{
    \textbf{(Left) Transferability of effort control  across different domains.} Despite training on math questions, we find that after AEC training, we can modulate the number of generated tokens and performance on different domains by varying the effort parameter $r$.
    Across all of the datasets we find that the number of generated tokens and accuracy increase overall with the effort parameter supplied in the prompt.
    \textbf{(Right) Effect of reward function and robustness to prompt}. We compare our AEC reward function to a modified reward function (Eq.~\ref{eq:len-penalty}) that penalizes the length of the generation regardless of problem difficulty. 
    For this experiment we use a modified prompt that specifies the effort through a ``points'' parameter $p$.
    After training with our AEC reward function (circles), we observe the ability to increase accuracy and the number of generated tokens by increasing $p$, highlighting that the dependence on the effort parameter does not depend on the specific prompt used. Training with a small length penalty ($\lambda=1e-4$) does not allow modulating the number of generated tokens or accuracy by varying the effort parameter $p$ (squares). Training with a large length penalty ($\lambda=1e-3$, triangles) leads to the ability to modulate the number of tokens and accuracy (within a narrow range), but results in significantly fewer generated tokens (less than 2000 for all datasets).
    }
    \label{fig:ood_and_effect-objective}
\end{figure}

We varied $p~\in\{1,2,3,4,5\}$, which corresponded to linearly varying our effort parameter $r \in [0.48, 0.8]$ 
in Eq.~\ref{eq:main-objective}.
In Fig.~\ref{fig:ood_and_effect-objective} (Right) we show that training with this prompt variant also leads to an overall increase in performance and number of generated tokens with increasing effort (now specified through ``points").
This shows that our sensitivity to the effort parameter is not dependent on the exact prompt used.

To ablate whether the dependency on the effort parameter depended on the specifics of the reward function used, we compared our reward function with a modified reward function that correspond to a length penalty denoted by $R_{\lambda}(y,y^*)$ where
\begin{equation}
\label{eq:len-penalty}
R_{\lambda}(y,y^*)=\mathbb{I}(y=y^*)- \frac{\lambda}{p} T_y.   
\end{equation}
In Fig.~\ref{fig:ood_and_effect-objective} (Right), we used either a small length penalty $\lambda=1e-4$ or a large length penalty $\lambda=1e-3$ and similarly varied $p~\in\{1,2,3,4,5\}$. When the length penalty is small with $\lambda=1e-4$ (shown in squares in the figure) while we observe a reduction in token lengths (and increase in accuracy compared to the baseline), we do not observe the ability to modulate the number of generated tokens though this parameter $p$. When we used a larger length penalty of $\lambda=1e-3$, we observed the ability to modulate the number of generated tokens and performance through the parameter $p$, though the number of tokens and performance was significantly reduced (triangles in plot). The training with a length penalty is therefore highly sensitive to the value of $\lambda$, whereas our approach is more robust as it depends on the statistics of the model during a rollout, without requiring this hyperparameter.

We also ablate the use of $T_{avg}$ in our AEC reward function in Eq.~{\ref{eq:main-objective}} (using our original effort prompt and settings). In particular we compare against training with the following maximum token constraint reward function
\begin{equation}
\label{eq:max_reward}
    R_{max}(y,y^*) = \mathbb{I}(y=y^*) \cdot \mathbb{I}(T_y < T_{max}),
\end{equation}
where $T_{max}$ is specified in the prompt and uniformly sampled in the range $T_{max} \sim U[100, 15000]$ per question. We find that training with the AEC reward function enables control over a larger range of tokens earlier in training
compared to using the $R_{max}$ reward function (Fig.~\ref{fig:app-effort_vs_max}) .

\subsection{Selecting effort level for a task: Conceptual Depth of the trained model}

We now examine how we can leverage our ability to control the effort level to select an optimal operating point for a model in a particular environment using a criterion that jointly accounts for performance and time, with a unit conversion factor $\lambda$ that is specific to the environment, task, user, and model. We call the compound criterion that jointly accounts for performance and time \emph{conceptual depth}, which is a property of the trained model, and is related to various notions of complexity that also account for the \emph{time} to produce the solution (\textit{e.g.,} Logical Depth \citep{bennett1988logical}, %
Levin Complexity \citep{li2019introduction}). %
The conceptual depth curve 
measures the tradeoff between error $E$ and time $T$ in a given environment, modulated by $\lambda_{\mathcal{D}}$:
\begin{equation}
\label{eq:conceptual-depth}
\kappa(T, \mathcal{D}, \lambda_{\mathcal{D}}) = E_\mathcal{D}(T) + \lambda_{\mathcal{D}}T,    
\end{equation}
where $\lambda_{\mathcal{D}}$ represents the cost of time as determined by the environment, $E_\mathcal{D}$ denotes the error rate on a dataset $\mathcal{D}$, and $T$ denotes the average length of the chain-of-thought when solving queries from the task. 
For a given environment, the minimum value of this curve measures the \textit{conceptual depth} of the model, and represents an optimal operating point.
Note $\kappa$ corresponds to $1 - J$; plotting this quantity across $T$ allows clearer scaling visualizations in a log-log plot. 

\begin{figure}[t]
    \centering
    \includegraphics[width=0.99\linewidth]{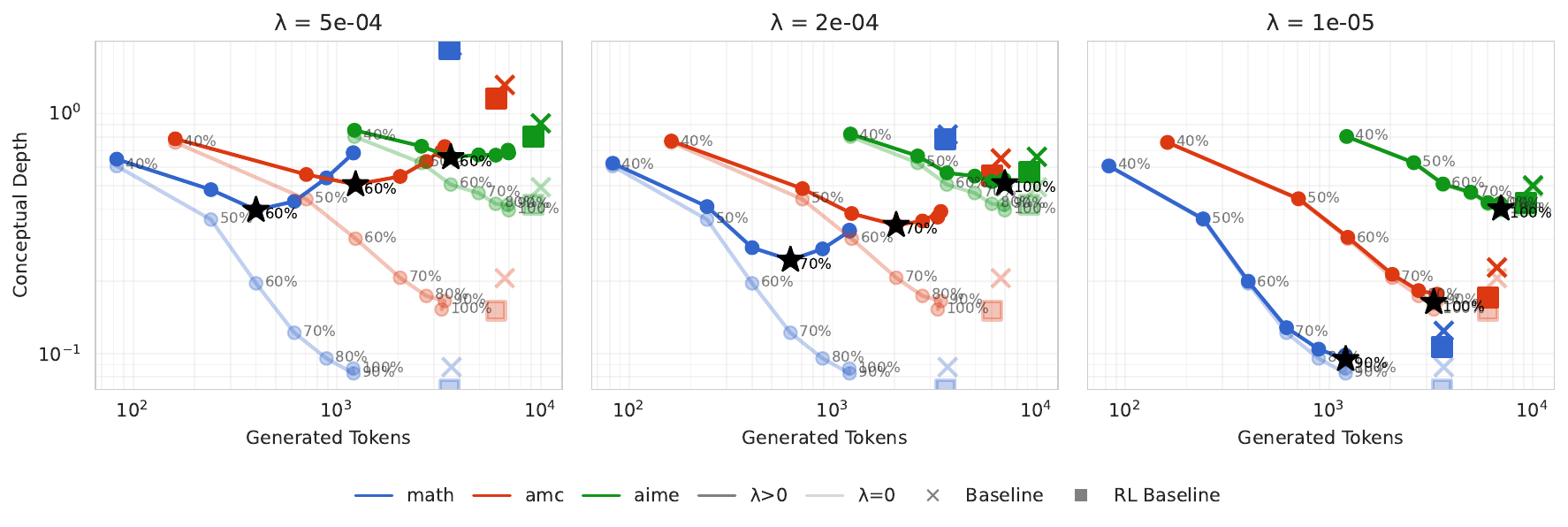}
    \vspace{-1em}
    \caption{\textbf{Conceptual Depth and task-specific effort selection.} A user or agent  should select the number of chain of thought tokens to balance the performance on the task and the cost incurred for achieving that performance. In math exams such as AMC (red) and AIME (green), the correct answer rendered after the time limit is worthless, and different exams have different time limits. When time and loss are compounded through the cost of time, the minimum value measures what we call the  {\em  conceptual depth} of the model. An ``intelligent'' agent should be able to choose this operating point, shown as black stars, depending on the specific task and query at hand. The error rate (where $\lambda=0$) is shown in a lighter shade. The plotted effort levels range from 40\% to 100\%. The optimal operating point can be specified by either the number of tokens or the effort level. For a given $\lambda$, the optimal effort level is similar across the different exams, even though the optimal number of tokens differs significantly depending on the particular exam and its time limit.}
    \vspace{-1em}
    \label{fig:app-conceptual-depth-math}
\end{figure}

To optimally specify the number of tokens/effort for a task given this objective, we should select the optimal chain-of-thought length $T^*$ such that
\begin{equation*}
    T^*(\mathcal{D}, \lambda_{\mathcal{D}}) = \arg \min_T \kappa(T, \mathcal{D}, \lambda_{\mathcal{D}}).
\end{equation*}
The minimum occurs when
 \[\frac{dE_\mathcal{D}(T)}{dT} = -\lambda_{\mathcal{D}}. \]
In Fig.~\ref{fig:app-conceptual-depth-math}, we plot the conceptual depth curves for various values of $\lambda_{\mathcal{D}}$ in math tasks of varying difficulty. 
We set $\lambda_{\mathcal{D}}$ to be inversely proportional to the time constraints in actual testing scenarios: $\lambda_{MATH} = \lambda$, $\lambda_{AMC} = \lambda/3$, $\lambda_{AIME} = \lambda/12$, reflecting that students receive 12 minutes per AIME problem and 3 minutes per AMC problem, while we approximate (as this is not a standardized exam) 1 minute per MATH500 problem.
We vary the value of $\lambda$ across the different panels while keeping the scaling across the datasets fixed.
We show that even though the number of tokens needed to minimize $\kappa$ varies significantly across the different math tasks, 
the effort level needed to minimize $\kappa$ is similar across the different tasks for a given value of $\lambda$. We show this is the case for various values of $\lambda$ (shown across the 3 panels). 
For larger values of $\lambda$ the models use a smaller effort level, and for smaller values of $\lambda$, the models use a larger effort level. %

The minimum of $\kappa$ occurs at similar effort levels across the different math datasets as $\frac{dE_\mathcal{D}(T)}{dT}$ evaluated at the different effort levels is inversely proportional to the (real) time allocated per question, which scales $\lambda_{\mathcal{D}}$ (Fig.~\ref{fig:delta_error_conceptual_depth}).
Controlling the effort therefore provides a simple and practical knob to select an optimal operating point across different tasks with different cost-of-time. %

\section{Conclusion}

We introduced AEC, a reinforcement learning method for enabling prompt-based control over the \emph{relative} amount of chain-of-thought tokens for reasoning models, while being adaptive to the difficulty of a problem. 
In addition to enabling control over the generation length, our approach enables a 1.9-$3.7$x reduction in chain of thought length at matched accuracy relative to the base model.

Our approach requires just a simple modification to traditional RL training. In this work we specified a single query-dependent parameter in the input during RL training and using a corresponding query-dependent reward function, which allowed inference-time control after training. Future work could extend our approach to specify multiple query-relevant parameters and using analogous example-dependent reward function to what we used. 
Additionally we applied our approach for a reasoning model -- future work could extend our training approach to multi-agent systems. 

\newpage
\bibliography{main}
\bibliographystyle{plainnat}

\newpage
\appendix
\label{sec:appendix}

\section{Qualitative Examples}

We show qualitative examples of the responses to an easy problem at effort level $50\%$ and $100\%$ in Fig.~\ref{fig:model_response_50} and Fig.~\ref{fig:model_response_100} respectively\footnote{We leveraged the LaTeX formatting in \citet{aggarwal2025l1} for presenting the generations of our effort control model.}. Note that for this question, while both solutions obtain the correct answer, the length of the solution is significantly shorter at $50\%$ effort level. We also show a generation for a more complicated AIME math question (Fig.~\ref{fig:model_response_50_aime}). 

\begin{figure}[h!]
    \centering
    \modelexamplecorrect{%
You have seven bags of gold coins. Each bag has the same number of gold coins. One day, you find a bag of 53 coins. You decide to redistribute the number of coins you have so that all eight bags you hold have the same number of coins. You successfully manage to redistribute all the coins, and you also note that you have more than 200 coins. What is the smallest number of coins you could have had before finding the bag of 53 coins?

Let's think step by step and output the final answer within \texttt{boxed\{\}}. Let's spend 50\% effort.
    }
    {
    50\%
    }
    {
     \begin{tcolorbox}[colback=gray!5,colframe=gray!75,title=Thinking Process,fonttitle=\small\bfseries]
      \small
    Let n be the original number per bag. Total before:7n +53. After adding 53, total is 8m. So $7n +53=8m$. Also $8m >200 => m>25$. So $7n=8m -53$. $8m -53$ must be divisible by7. 8m $\equiv$53 mod7. 53 mod7=4, 8$\equiv$1 mod7, so m$\equiv$4 mod7. Smallest m$>$25 with m$\equiv$4 mod7: 26? 26 mod7=5. 27 mod7=6. 28 mod7=0. 29 mod7=1. 30 mod7=2. 31 mod7=3. 32 mod7=4. So m=32. Total coins=8*32=256. Before:256-53=203.
    \end{tcolorbox}
    \boxed{203}
    }    
    {
    203
    }
    {
    224
    }

  \caption{Example model response with effort level 50\% for a problem from MATH500. 224 tokens were generated.}
  \label{fig:model_response_50}
\end{figure}

\begin{figure}[t!]
    \centering
    \modelexamplecorrect{%
You have seven bags of gold coins. Each bag has the same number of gold coins. One day, you find a bag of 53 coins. You decide to redistribute the number of coins you have so that all eight bags you hold have the same number of coins. You successfully manage to redistribute all the coins, and you also note that you have more than 200 coins. What is the smallest number of coins you could have had before finding the bag of 53 coins?

Let's think step by step and output the final answer within \texttt{boxed\{\}}. Let's spend 100\% effort.
    }
    {
    100\%
    }
    {
     \begin{tcolorbox}[colback=gray!5,colframe=gray!75,title=Thinking Process,fonttitle=\small\bfseries]
      \small
    Let me try to figure out this problem step by step. So, the setup is that I have seven bags of gold coins, each containing the same number of coins. Then, I find an additional bag with 53 coins, making it eight bags in total. I redistribute all the coins so that each of the eight bags has the same number of coins. After redistribution, I note that I have more than 200 coins in total. I need to find the smallest number of coins I could have had before finding the 53-coin bag.

Okay, let me break this down. Let's denote the number of coins in each of the original seven bags as 'n'. So, the total number of coins I had initially is 7n. Then, I found a bag with 53 coins, so the total number of coins after finding it becomes 7n + 53.

Now, I redistribute all these coins equally among eight bags. Let's denote the number of coins in each of the eight bags after redistribution as 'm'. So, we have 8m = 7n + 53. This equation tells me that 7n + 53 must be divisible by 8 because m has to be an integer.

[Thinking process truncated for visualization ...]

Therefore, the smallest n is when k=3:

n=8*3 +5=24 +5=29.

So, n=29.

Let me verify this. If n=29, then the initial total is 7*29=203. Adding 53 gives 203 +53=256.

256 divided by8 is 32, which is an integer. So, that works. Also, 256 is more than 200, as required.

Therefore, the smallest number of coins I could have had before finding the bag of 53 coins is 203.
    \end{tcolorbox}
    The smallest number of coins you could have had before finding the bag of 53 coins is \boxed{203}.
    }
    {
    203
    }
    {
    1183
    }

  \caption{Example model response with effort level 100\% for a problem from MATH500. 1183 tokens were generated.}
  \label{fig:model_response_100}
\end{figure}

\begin{figure}[t!]
    \centering
    \modelexamplecorrect{%
Each vertex of a regular octagon is independently colored either red or blue with equal probability. The probability that the octagon can then be rotated so that all of the blue vertices end up at positions where there were originally red vertices is $\tfrac{m}{n}$, where $m$ and $n$ are relatively prime positive integers. What is $m+n$?

Let's think step by step and output the final answer within \texttt{boxed\{\}}. Let's spend 50\% effort.
    }
    {
    50\%
    }
    {
     \begin{tcolorbox}[colback=gray!5,colframe=gray!75,title=Thinking Process,fonttitle=\small\bfseries]
      \small

The probability is the number of colorings fixed by some rotation divided by the total number of colorings. But maybe it's easier to use Burnside's lemma. However, the problem is about rotations where blue vertices end up on originally red positions. So we need colorings where the blue vertices are a subset of the red vertices under some rotation. This is equivalent to the set of blue vertices being a union of some orbits under rotation. Since the octagon is regular, the group is cyclic of order 8. The possible rotations that could fix the condition are those by multiples of 45 degrees. For each divisor d of 8, we look at rotations by 360/d degrees. The number of colorings fixed by rotation by 360/d degrees is $2^{8/d}$. But I might be mixing things up. Alternatively,

[Thinking process truncated for visualization ...]

But wait, no, this counts the number of colorings fixed by each rotation, but overlapping colorings are counted multiple times.

So to get the actual number, inclusion-exclusion is needed, but it's complicated.

Alternatively, since the problem is about probability, and the total colorings is 256, and the average number of fixed colorings is 290 /8=36.25, which is not an integer. So perhaps this approach is wrong.

Alternatively, maybe the number is 36, as computed before, but 36/256=9/64, which is m/n=9/64, m+n=73.

But I'm not sure.
    \end{tcolorbox}
The probability is $\frac{9}{64}$, so $m + n = 9 + 64 = 73$.

\boxed{73}
    }
    {
    371
    }
    {
    3503
    }

  \caption{Example model response with effort level 50\% for a problem from AIME. 3503 tokens were generated.}
  \label{fig:model_response_50_aime}
\end{figure}

\FloatBarrier

\section{Additional Results}

\begin{figure}[h]
    \centering
    \includegraphics[width=0.98\linewidth]{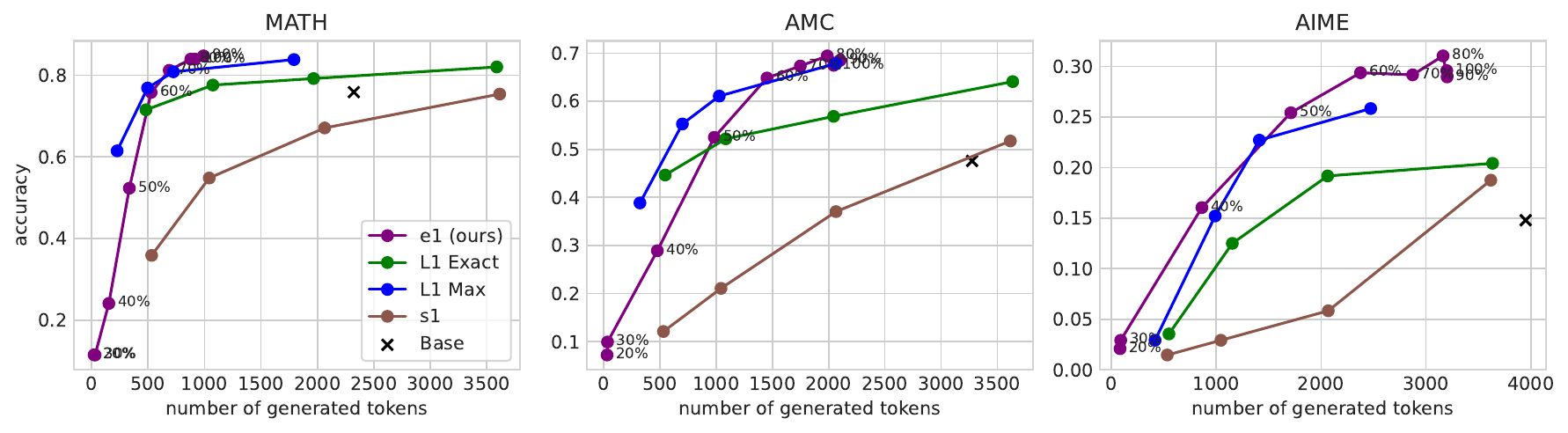}
    \caption{Comparison against other token-based control approaches separated per dataset. Same setup as Fig.~\ref{fig:l1-comparison} (Left).}
    \label{fig:app-l1-comparison}
\end{figure}

\begin{figure}[h]
    \centering
    \includegraphics[width=0.98\linewidth]{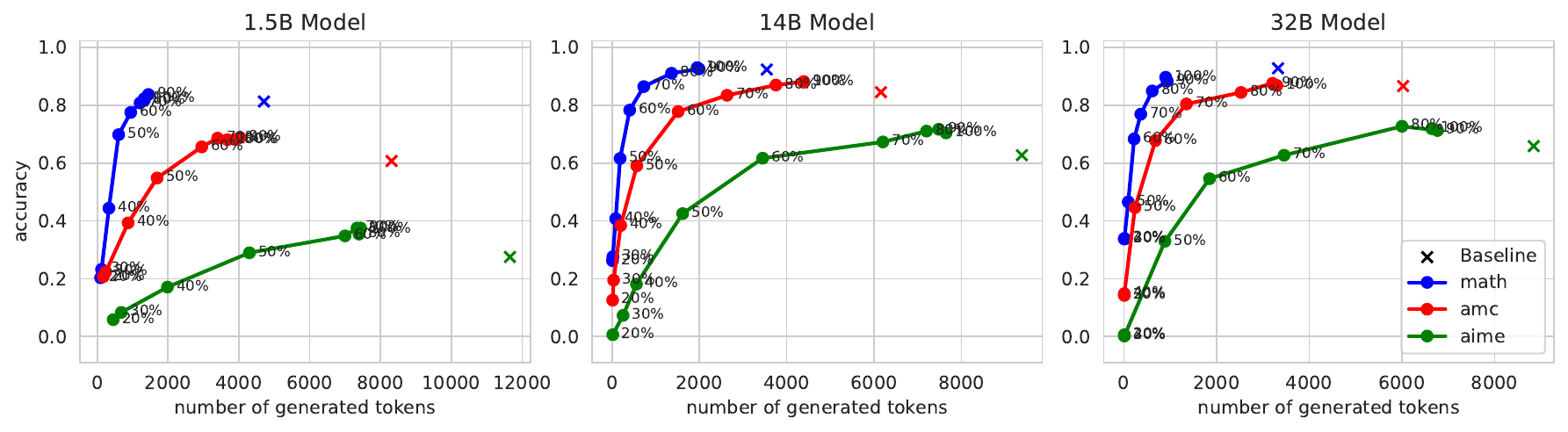}
    \caption{Same as Fig.~\ref{fig:main_math_and_ood} (Left) for various model sizes.
    Together with Fig.~\ref{fig:main_math_and_ood} (Left), these were aggregated to form Fig.~\ref{fig:main_math_and_ood} (Right). 
    }
    \label{fig:app-model-sizes}
\end{figure}

\FloatBarrier
\begin{table}[ht]
\centering
\caption{Per-dataset reduction in generated tokens at matched accuracy relative to the base model, computed from Fig.~\ref{fig:main_math_and_ood} and Fig.~\ref{fig:app-model-sizes}. Token reduction is computed by comparing the number of generated tokens at the lowest effort level that matches the target accuracy, with linear interpolation between adjacent effort levels. Averages are computed over the datasets in which the base model's accuracy is reached (on MATH500 for the 32B it is not reached; our model reaches $3$ percentage points below it at maximum effort while using 3.7$\times$ fewer tokens).}
\label{tab:reductions}
\begin{tabular}{lcccc}
\toprule
& \multicolumn{4}{c}{Token reduction at matched accuracy ($\times$)} \\
\cmidrule(lr){2-5}
Model & MATH500 & AMC & AIME & Avg. \\
\midrule
1.5B & 3.7 & 3.5 & 2.9 & 3.4 \\
7B   & 3.4    & 3.2 & 2.5 & 3.0 \\
14B  & 1.9    & 2.1 & 2.4 & 2.1 \\
32B  & --- & 2.0 & 2.1 & 2.0 \\
\bottomrule
\end{tabular}

\vspace{0.5em}
\end{table}

\begin{figure}[h]
    \centering
    \includegraphics[width=0.4\linewidth]{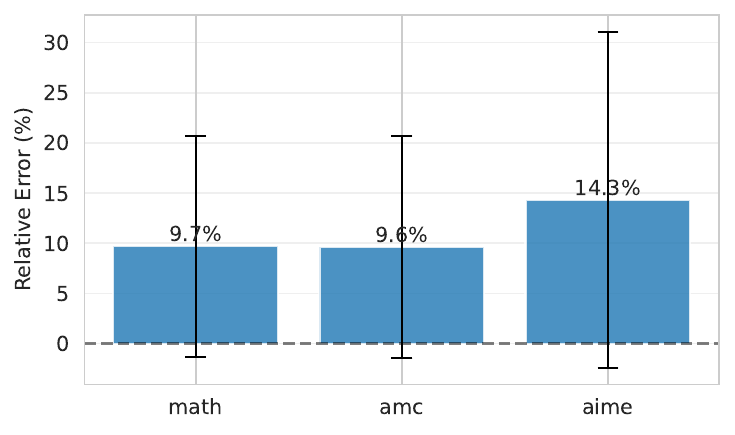}
    \caption{Calibration of percentage effort across datasets. We plot the average error (and standard deviation) across effort values from Fig.~\ref{fig:dynamics-calibration} (Right) for each dataset.}
    \label{fig:app-calibration}
\end{figure}

\begin{figure}[h]
    \centering
    \includegraphics[width=0.6\linewidth]{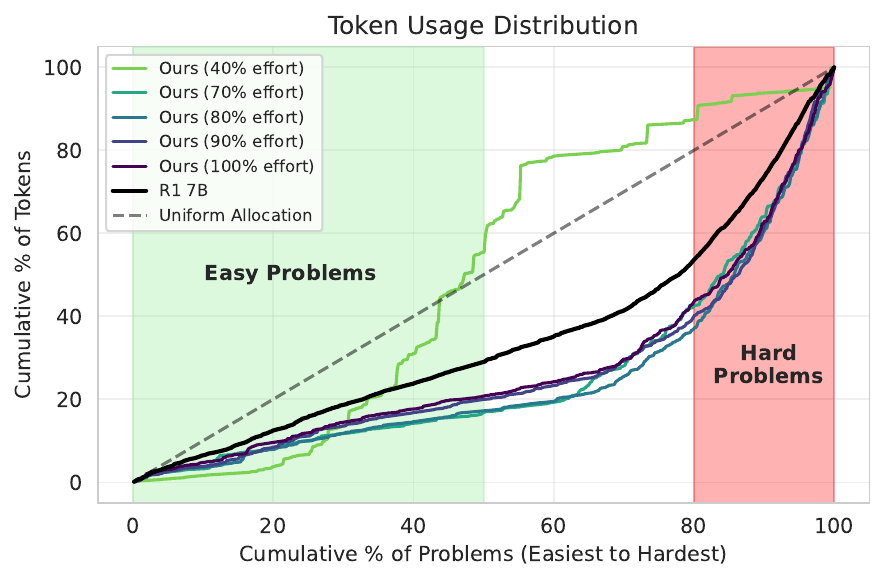}
    \caption{\textbf{Adaptation to query difficulty for 7B model, using R1 Distilled Qwen as Base Model.} In addition to providing control over the number of tokens, we show that training with our approach leads to smaller token allocation to easier problems, and larger token allocation to more difficult problems. We show results for various values of $r$. For values of $r \geq 0.7$, we observe fewer token allocation to easier problems, but not for low percentage effort when the total number of generated tokens and accuracies were significantly lower.
    }
    \label{fig:difficulty-adaptation}
\end{figure}

\begin{figure}[h]
    \centering
    \includegraphics[width=0.4\linewidth]{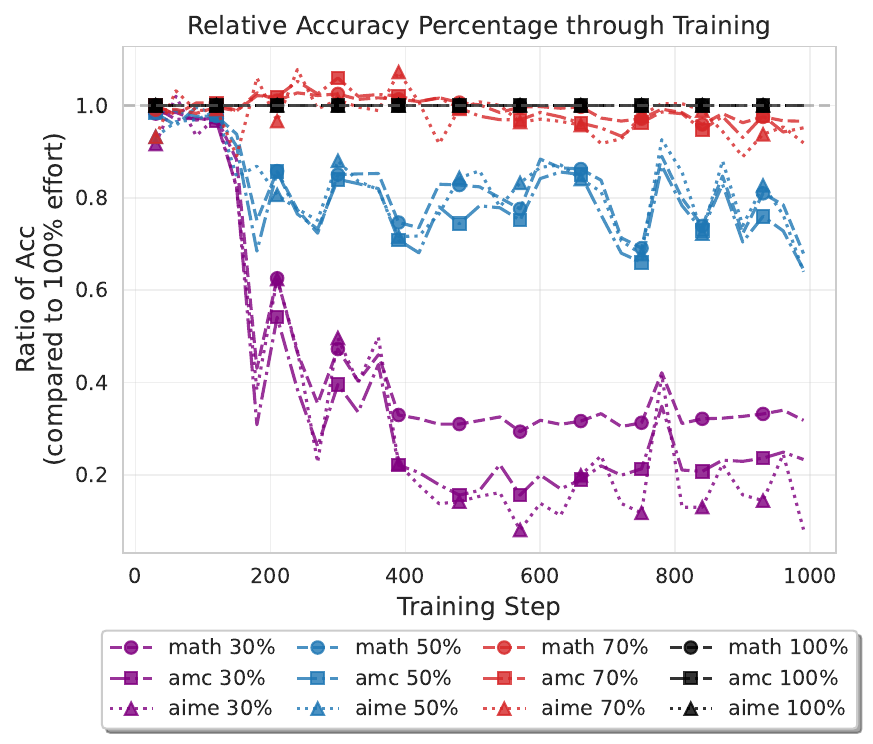}
    \includegraphics[width=0.4\linewidth]{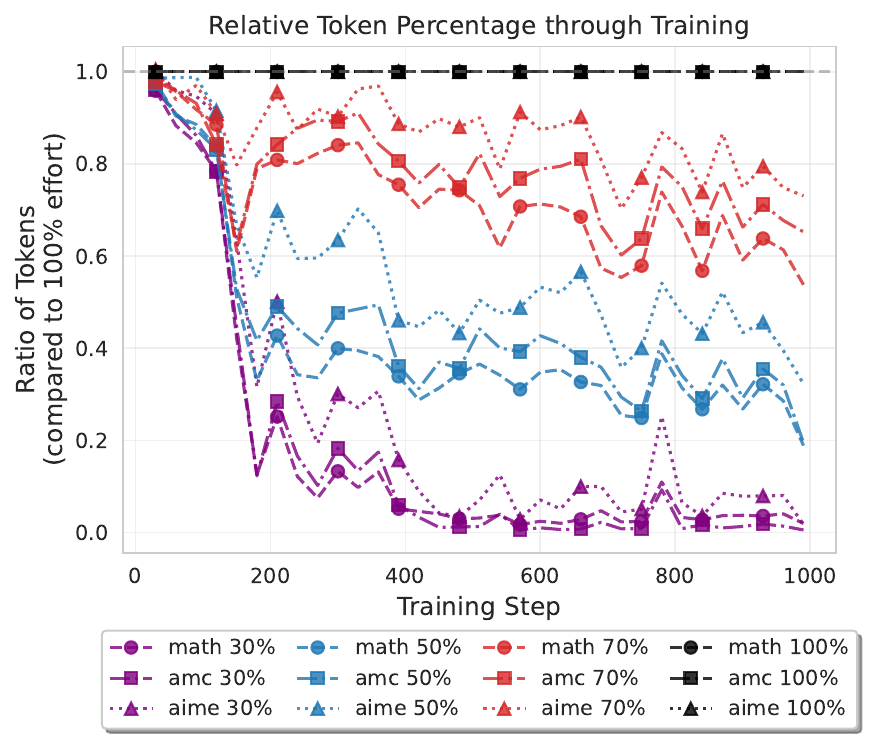}
    \caption{
   \textbf{Relative accuracy and relative token length through training.} 
    We evaluate the relative accuracy and relative number of tokens through training by comparing relative to $100\%$ effort. Despite problems being of varying difficulty and taking varying length, we find that the relative accuracy across datasets remains roughly constant over training for a fixed effort level, as the curves are overlapping. 
    We observe a similar phenomenon when analyzing the ratio of number of tokens.}
    \label{fig:relative-toks-acc}
\end{figure}

\begin{figure}[h]
    \centering
    \includegraphics[width=0.97\linewidth]{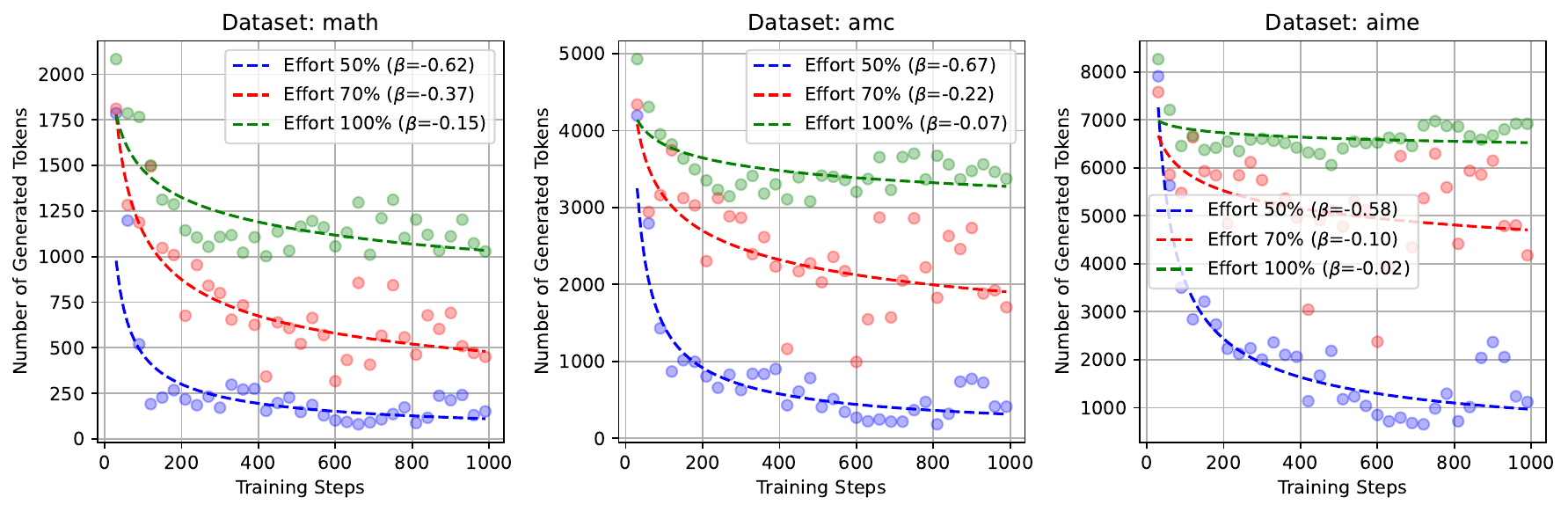}
    \includegraphics[width=0.97\linewidth]{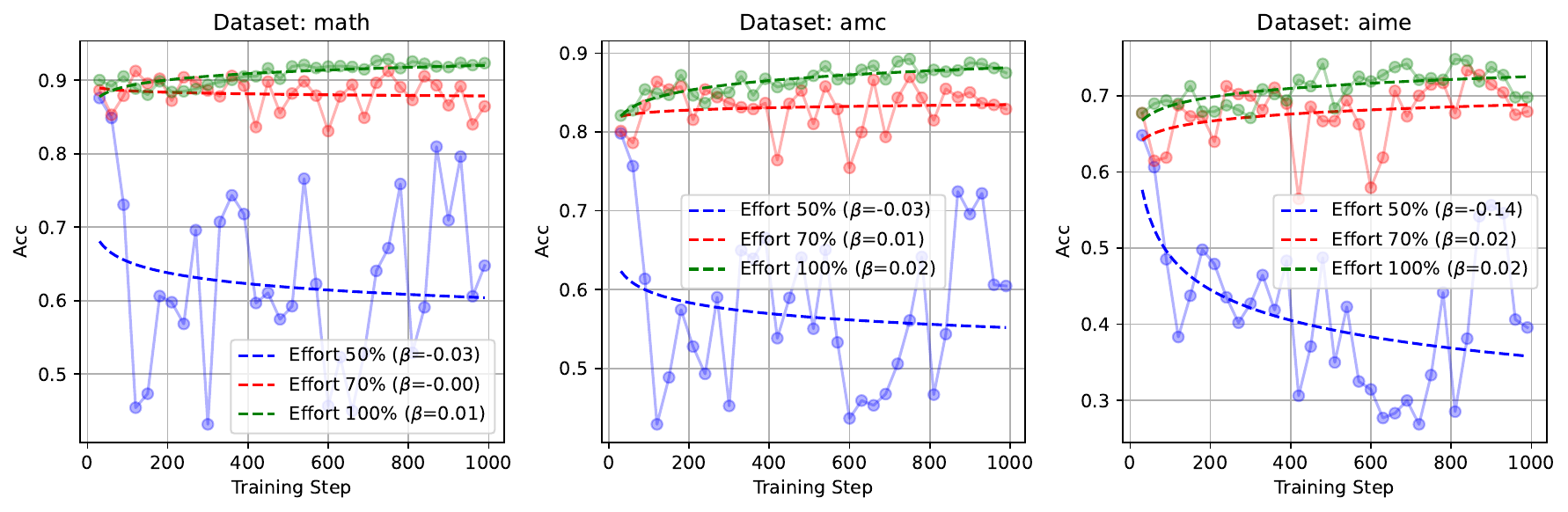}
    \caption{\textbf{Learning dynamics of AEC training for 32B model.} 
We plot the accuracy and number of tokens as a function of training step for the three math evaluation datasets. Dashed lines indicate a power-law fit with exponent $\beta$. 
    }
    \label{fig:learning-dynamics-32b}
\end{figure}

\begin{figure}[h]
    \centering
    \includegraphics[width=0.97\linewidth]{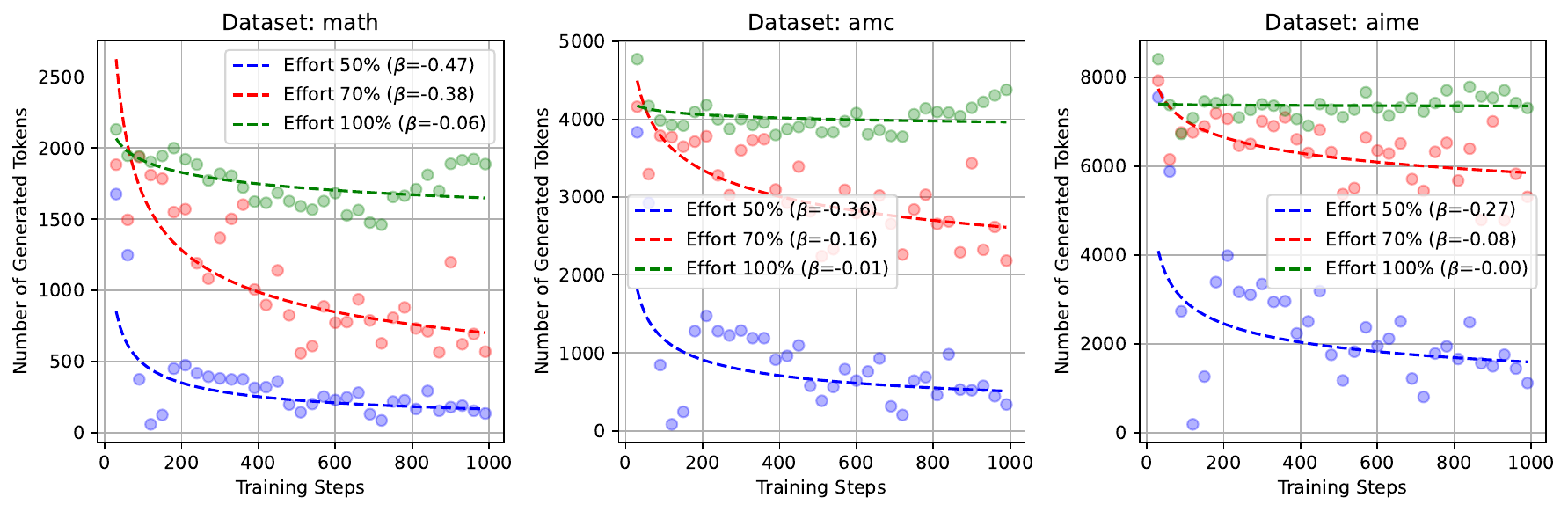}
    \includegraphics[width=0.97\linewidth]{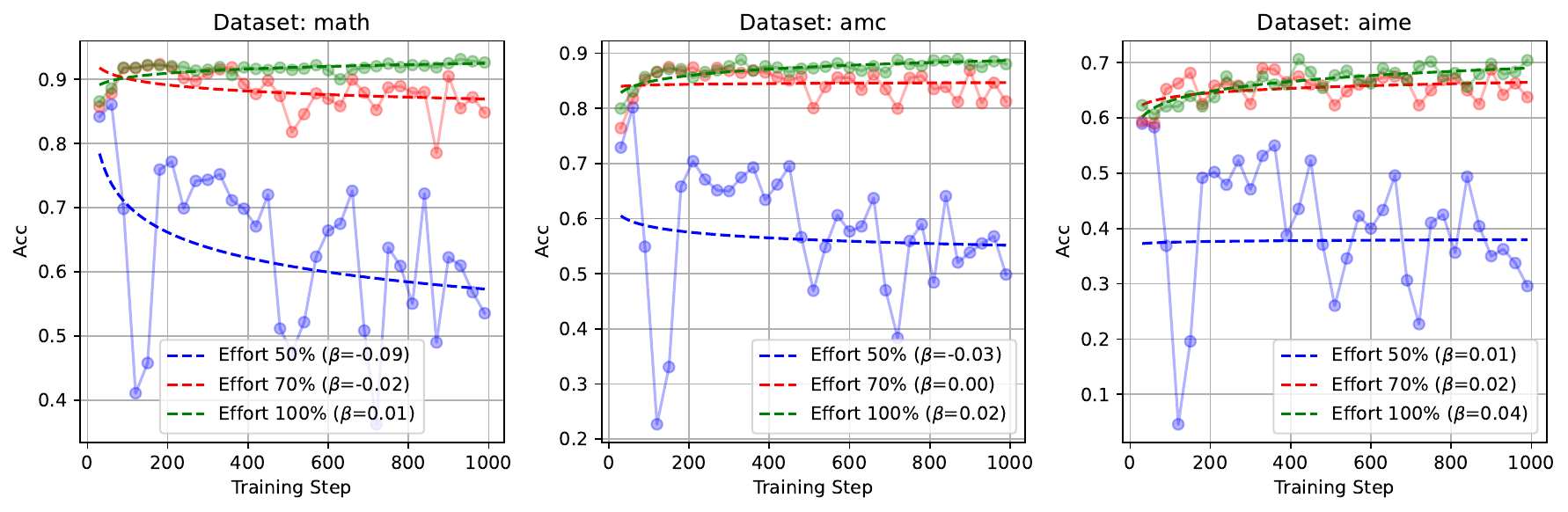}
    \caption{\textbf{Learning dynamics of AEC training for 14B model.} 
We plot the accuracy and number of tokens as a function of training step for the three math evaluation datasets. 
Dashed lines indicate a power-law fit with exponent $\beta$. 
    }
    \label{fig:learning-dynamics-14b}
\end{figure}

\begin{figure}[h]
    \centering
    \includegraphics[width=0.97\linewidth]{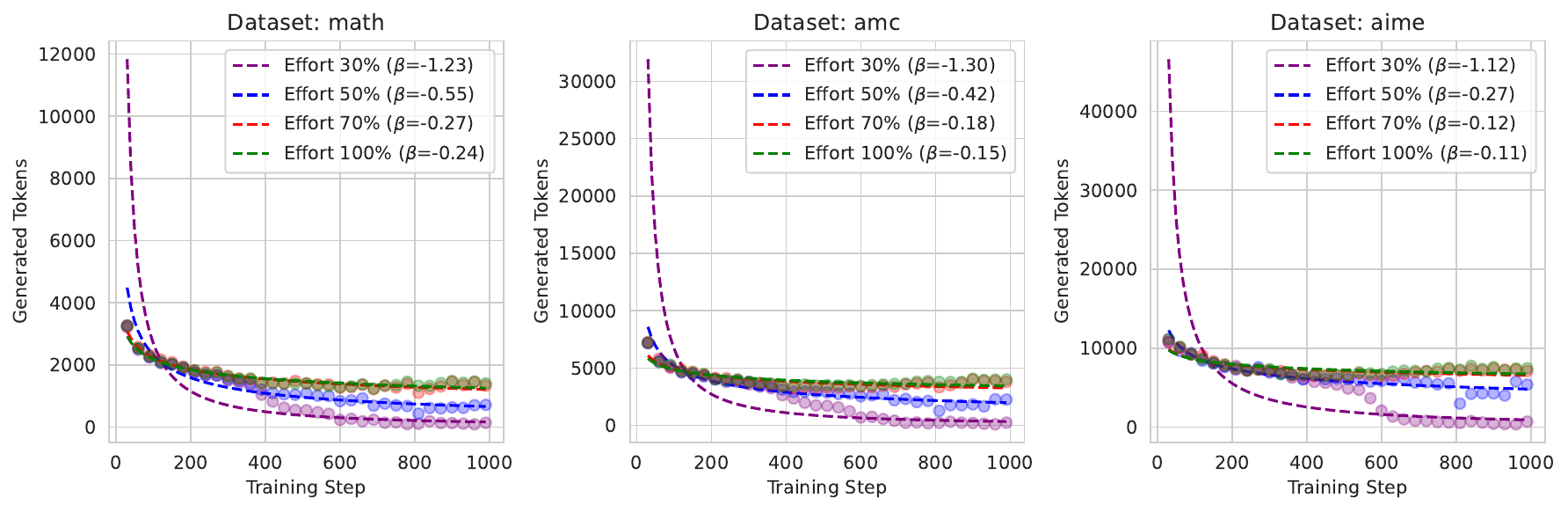}
    \includegraphics[width=0.97\linewidth]{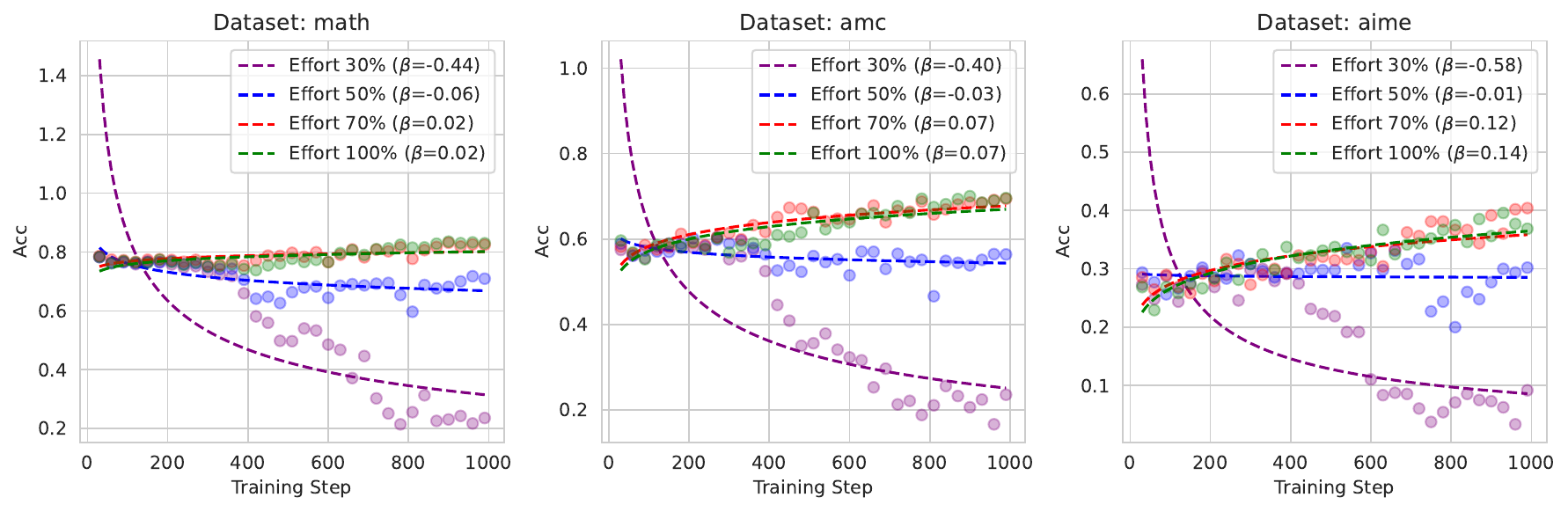}
    \caption{\textbf{Learning dynamics of AEC training for 1.5B model.} 
We first plot the accuracy and number of tokens as a function of training step for the three math evaluation datasets. Dashed lines indicate a power-law fit with exponent $\beta$.
    }
    \label{fig:learning-dynamics-1.5b}
\end{figure}

\begin{figure}
    \centering
    \includegraphics[width=0.99\linewidth]{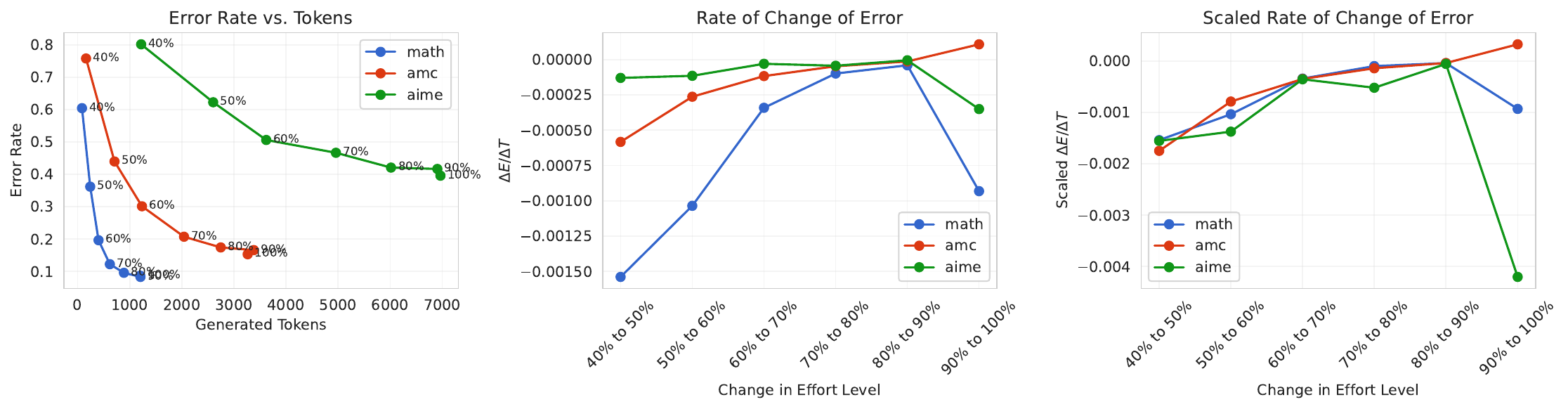}
    \caption{\textbf{(Left)} We plot the error rate across generated tokens for effort levels $40\%$ to $100\%$.
    \textbf{(Center)} We compute the change in error over the change in tokens across the different effort levels (dashed lines). \textbf{(Right)} When rescaling these curves by the time allocated to problems from these datasets (12 minutes for AIME, 3 minutes for AMC, and we approximate 1 minute for MATH500), we observe the rescaled curves are approximately overlapping (solid lines). This means that the change in error rate $\frac{dE_D(T)}{dT}$ for different datasets $\mathcal{D}$ is inversely proportional to the real time allocated per question at different effort levels.
    }
    \label{fig:delta_error_conceptual_depth}
\end{figure}

\begin{figure}
    \centering
    \includegraphics[width=0.5\linewidth]{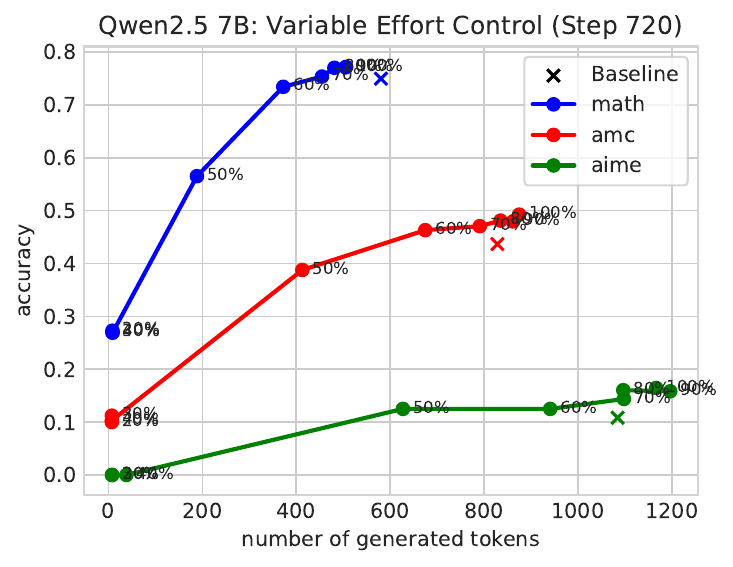}
    \caption{To evaluate whether AEC depends on an R1-distilled initialization, we trained Qwen 2.5-7B-Instruct using AEC (with a maximum response length of 4K tokens). After training, we observe that accuracy and token usage increase overall with effort level, analogous to our main results with R1-distilled models. 
    Results are shown at step 720.}
    \label{fig:app-effect-base}
\end{figure}

\begin{figure}
    \centering
    \includegraphics[width=0.32\linewidth]{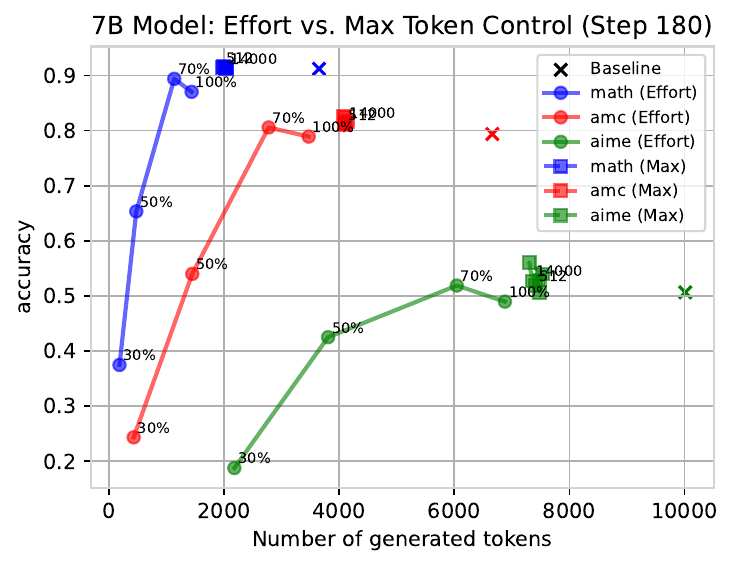}
    \includegraphics[width=0.32\linewidth]{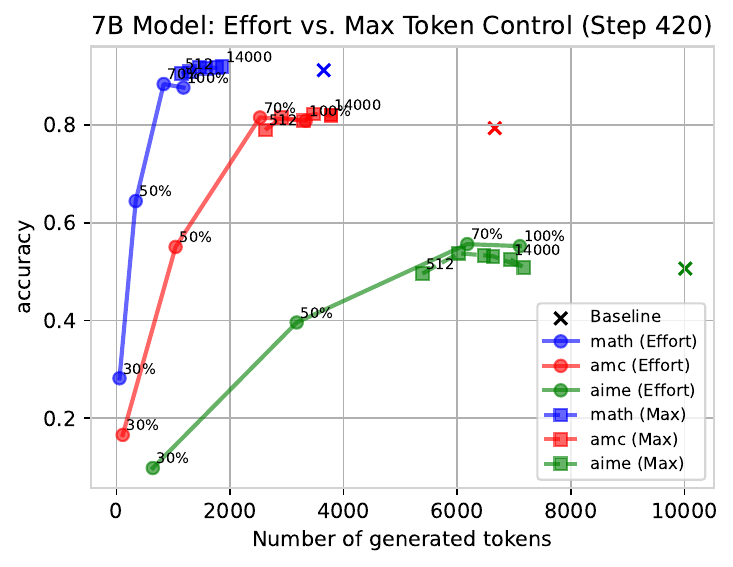}
    \includegraphics[width=0.32\linewidth]{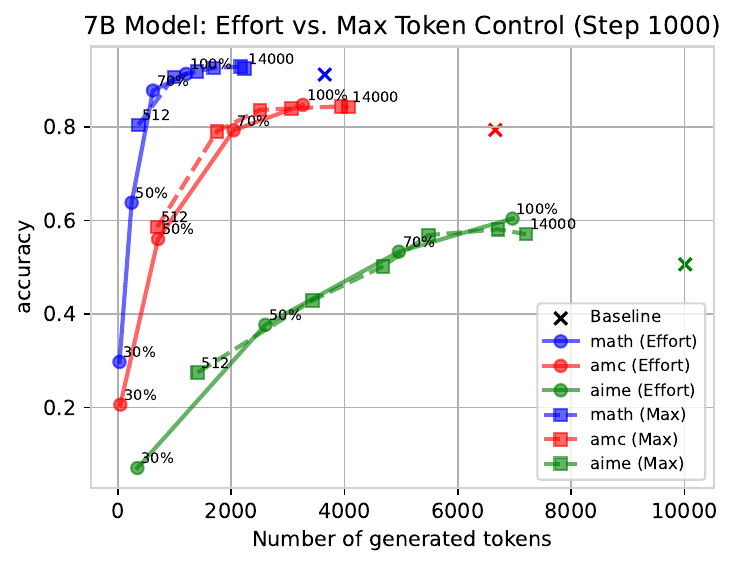}
    \caption{We compare training with the AEC reward function against using a maximum token-based reward function, where the maximum number of tokens is also specified in the prompt (Eq.~\ref{eq:max_reward}). We find the ability to control the number of tokens emerges earlier in training using our AEC objective (by step 180, left plot). Results are shown at steps 180 (Left), 420 (Center) and 1000 (Right). 
    }
    \label{fig:app-effort_vs_max}
\end{figure}

\end{document}